\documentclass[11pt,a4paper]{article}
\usepackage[]{acl}
\usepackage{times}
\usepackage[utf8]{inputenc}
\usepackage[main=english,bidi=default]{babel}
\usepackage{devanagari}
\usepackage{amsmath}

\usepackage{multirow}

\usepackage{microtype}

\title{Socially Aware Bias Measurements for Hindi Language Representations}

\author{Vijit Malik$^{1}$\thanks{~~Work done while interning at UCLA.} \quad \textbf{Sunipa Dev}$^{2}$ \quad
\textbf{Akihiro Nishi}$^{2}$ \quad
\textbf{Nanyun Peng}$^{2}$  \quad \textbf{Kai-Wei Chang}$^{2}$\\

 $^{1}$Indian Institute of Technology Kanpur (IIT-K) \\
 $^{2}$University of California, Los Angeles\\
\texttt{\{vijitvm\}@iitk.ac.in} \\ 
\texttt{\{akihironishi\}@ucla.edu}\\
\texttt{\{sunipa, violetpeng, kwchang\}@cs.ucla.edu}
}

\date{}

\begin{document}
\maketitle
\begin{abstract}
\textcolor{red}{\emph{Trigger warning:} This paper contains examples of stereotypes and other harms that could
be offensive and triggering to individuals.}

Language representations are efficient tools used across NLP applications, but they are strife with encoded societal biases. These biases are studied extensively, but with a primary focus on English language representations and biases common in the context of Western society. In this work, we investigate biases present in Hindi language representations with focuses on caste and religion-associated biases. We demonstrate how biases are unique to specific language representations based on the history and culture of the region they are widely spoken in, and  how the same societal bias (such as binary gender-associated biases) is encoded by different words and text spans across languages. 
The discoveries of our work highlight the necessity of culture awareness and linguistic artifacts when modeling language representations, in order to better understand the encoded biases.
\end{abstract}

\section{Introduction}
\vspace{-0.2cm}
Language models and representations \cite{pennington2014glove, bojanowski2017enriching, devlin-etal-2019-bert} are commonly used across the world in a variety of applications including machine translation \cite{kunchukuttan2017iit}, information retrieval \cite{rao2018eventxtract}, chatbots \cite{Bhagwat2019ReviewOC}, sentiment classification \cite{kumar2019bhaav} and more. 
However, it is also known that these representations capture and propagate societal biases including gender~\cite{NIPS2016_a486cd07}, race~\cite{caliskan2017semantics,manzini-etal-2019-black}, and nationality~\cite{dev2019attenuating} related stereotypes. 
This bias is present across representations for different languages. 
Each language reflects the culture and history of regions they are used popularly in, and as we go from one language to another, the notion of bias, and the types of societal biases change accordingly. This key difference however, is not reflected in the effort made towards detecting, identifying, and mitigating biases in  language representations, with the majority of efforts predominantly in English language and in the context of Western society~\cite{NIPS2016_a486cd07,nangia-etal-2020-crows}. 
Some recent work tackle the challenges of societal biases in language representations coming from various cultures and languages such as Arabic, French, Spanish, Chinese, and German~\cite{lauscher-etal-2020-araweat,chavez-mulsa-spanakis-2020-evaluating,kurpicz2020cultural,zhou-etal-2019-examining,zhao-etal-2020-gender,liang-etal-2020-monolingual}. Additionally, \citet{ghosh2021detecting} and \citet{Sambasivan2020NonportabilityOA} explore biases and algorithmic fairness concerning non-western contexts in machine learning and 
~\citet{10.1145/3377713.3377792} focus on fairness in language technologies for Indic society by investigating binary gender associated bias in Hindi language representations. However, it is unclear if the representations capture other biases that are \emph{distinct} to the Indian society and can cause harm, such as caste and religion related biases. 

In this work, we make three main contributions: (i) with a careful study of the social and cultural composition of the Indian society, we highlight and devise measures to detect and quantify different societal biases present in Hindi language representations, and show gender, caste and religion bias in the language;
(ii) we discuss the gendered nature of Hindi and the implications it has on the bias detection techniques for gender bias, highlighting the importance of leveraging linguistic knowledge when developing bias detection methods; and
(iii) we demonstrate how the translation of word lists or one-to-one mapping of bias measures across languages is insufficient to detect biases meaningfully, indicating how bias measurement methods cannot be directly adapted from one language to another. 
Even when detecting the same societal types of biases in a different language, translations of the words from English into the target language whose representations are being evaluated, does not suffice as the words may not exist, nor be commonly used or associated with the same sentiment or bias. Sentiments (good, bad, evil, etc) can also be encoded by distinct sets of objects and words (see Section \ref{sec:investigating} for discussion). 

All these discoveries call for socio-cultural awareness, and attention to the differences in language structures and grammars in multilingual fairness studies. 
We hope this work can shed lights for future studies in these directions\footnote{Code is available at \href{https://github.com/vijit-m/SocHindi}{https://github.com/vijit-m/SocHindi}}. 

\section{Language and Culture Considerations}
\label{sec: language and culture}
\vspace{-2mm}
The perception and interpretation of biases are sensitive to societal construct, socio-cultural structures, and historical compositions of different regions of the world \cite{doi:10.1177/1028315306291538}. 
Since languages are culture and region-specific, there is a requirement to study the socio-cultural differences when trying to understand the biases and potential harms of language representations globally. 

Consider the example of the Hindi language where along with gender bias \cite{amutha2017roots}, other biases like the ones based on caste and religion are also pervasive. Caste is unique to the culture in the Indian peninsula and is not usually considered when analyzing biases in Western languages 
but remnants of caste based stereotypes are still prolific in the modern Hindi literature \cite{GUPTA2021}. Similarly, region and culture-specific biases also are present with respect to religion, occupation, or location \cite{thorat2007caste}.  

Additionally, there are several key linguistic differences between English and Hindi languages such as pronouns which in Hindi do not indicate gender unlike in English. Instead, gender may be indicated by adjectives or verbs (Section \ref{section:gender_in_hindi}), thus requiring distinct strategies for gender bias detection.
Further, the word order in Hindi is distinct from that of English, and is similar to Japanese, Korean, Mongolian and Turkish\footnote{Word order refers to positioning of subject, verb, and object in a sentence.}. Unlike English, which has fixed word order, Hindi does not has fixed word order. 

\subsection{Gender in Hindi}
\vspace{-0.1cm}
\label{section:gender_in_hindi}
The syntactical use of gender in Hindi is layered and distinct from English~\cite{inbook} in different ways. These differences are reflected in the structure and composition of text, and is essential when interpreting the ways biases are likely to be encoded.

\paragraph{Gendered verbs:} Verbs in Hindi can be gendered depending upon the tense of the sentence. In case of past tense and the perfect tenses 
which are built upon the past tense, the gender of the verb is determined by the gender of the subject. For example, `went' is `gaya' if male and `gayi' if female. 
    
\paragraph{Gendered Adjectives:} Adjectives in Hindi can also be gendered. However, not all adjectives change form according to the gender of the subject. For example, the adjective `gharelu' (Domestic) in Hindi is used the same whether a man is domestic or a woman is domestic, but adjectives like `good' is `achha' if male and `acchhi' if female.
    
\paragraph{Gendered titles:} Some titles for entities in Hindi can be gendered. 
For example: `teacher' is `adhyapak' if male and `adhyapika' if female.
    
\paragraph{Gendered inanimate nouns:} Instantiations of grammatical gender for inanimate nouns when used in a sentence is an important aspect of Hindi language. Note that these instantiations also depend upon the dialect spoken \cite{inbook}. The word `dahi' (yogurt) is assigned feminine forms of verbs and adjectives in western dialects and masculine forms in Eastern dialects of Hindi.

\subsection{Caste, Religion, and Occupation Biases}
\vspace{-0.2mm}
Historically, discrimination based on attributes of caste, religion, and occupation has been prominent in India \cite{banerjee1985caste, deshpande2010history}. While illegal, some of these biases are still common and as a result, language resources reflect the same.  
Caste is a form of social stratification unique to Hinduism and Indian society. It involves the concept of hereditary transmission of the style of life in terms of occupation, status, and social interaction, and is commonly associated strongly with last names of persons
\footnote{\href{https://en.wikipedia.org/wiki/Caste}{https://en.wikipedia.org/wiki/Caste}}. 
Consequently, it is associated with strong biases pertaining to purity, goodness, intelligence, etc. of individuals, 
which is reflected commonly in Hindi corpora. 
Despite being a secular nation, due to historical clashes, there are biases in India against the relatively minority religious population practicing Islam as opposed to the majority religion of Hinduism. These biases are associated with perceptions (good, bad, evil, etc.) of the different populations. Although biases against other religions are also present, we especially focus upon Islam and Hinduism since these are the most prominent. 
Like caste, some last names are highly associated with religion in India,  and can serve as a proxy 
for studying the bias. 
In addition to these biases, the gaps between rural and urban India in terms of the education and poverty has led to 
a discrepancy of perception (positive versus negative) between urban and rural occupations.

\section{Measuring Biases in Hindi language Representations}

\label{sec:investigating}


To quantify bias in English embeddings,  \citet{caliskan2017semantics} propose the Word Embedding Association Test (WEAT), which measures the stereotypical association between two sets of target concepts 
and attributes 
(See Appendix \ref{app: weat}), where 
a larger WEAT score indicates a larger bias.
\citet{may-etal-2019-measuring} propose SEAT (Sentence Embedding Association Test) for measuring bias in sentence encoders. Similar to word lists for WEAT tests, SEAT comprises of sentences in which each sentence is is a semantically neutral template which are completed with target words related to protected attributes and associated stereotypes. This puts focus on target words on which bias is to be measured.


\subsection{Gender Bias}

For evaluating binary gender-associated bias, we create WEAT tests \cite{caliskan2017semantics} in two ways in Hindi, by creating (i) \textit{Translated} word lists, and (ii) \textit{Language-Specific} word lists. 

For the \textit{Translated}\footnote{ Google-Translate API used to obtain translations.} test, we directly translate each individual word in each test for career \& family, arts \& mathematics and arts \& sciences tests in \cite{caliskan2017semantics}. Note that direct translations of some words like `Shakespeare' and `NASA' in Arts and Science lists, are not accurate, have ambiguous spellings, or are not populus in the literature. 
Also, some words from English like `cousin' do not have a corresponding word in Hindi.
Next, we develop a set of socially-aware \textit{Language-specific} tests, where we curate word lists (both attribute and target) (Appendix \ref{app:wordlists}) based on popular word usage in Hindi and their associations, 
and word frequencies in Hindi Wikipedia text. 

Table \ref{tab:trans-vs-crafted} highlights how translated sets capture lesser bias as compared to the WEAT tests specifically created for Hindi. 
In particular, the WEAT translated test for binary gender and science versus arts captures significantly low bias, unlike what is prevalent in the society as well as associated text and representations \cite{KHADILKAR2021100409,Madaan2018AnalyzeDA,articlepundir}. This, in turn, emphasizes the importance of creating language-specific tests.


\begin{table*}
\small
\centering
\begin{tabular}{|c|c|cc|cc|}
\hline
\multirow{2}{*}{\textbf{Attribute}}       & \multirow{2}{*}{\textbf{Description}}                       & \multicolumn{2}{c|}{\textbf{WEAT (GloVe)}}                                   & \multicolumn{2}{c|}{\textbf{SEAT (GloVe)}}                                   \\ \cline{3-6} 
                                          &                                                             & \textbf{Translated}              & \textbf{Lang-Specific}                          & \textbf{Translated}              & \textbf{Lang-Specific}                          \\ \hline
\multirow{2}{*}{\textbf{Gender}}          & maths, arts vs male, female                                 & 0.94 (0.02)                      & \textbf{1.12 (0.01)}                      & 0.87 (0.00)                      & \textbf{1.14 (0.00)}                     \\
                                          & science, arts vs male, female                               & 0.27 (0.31)                      & \textbf{1.13 (0.02)}                      & 0.18 (0.17)                      & \textbf{1.03 (0.00)}                      \\ \hline
\textbf{Caste}                            & adjectives vs caste                                         & 0.72 (0.00)                      & \textbf{1.52 (0.00)}                      & 0.74 (0.00)                      & \textbf{1.40 (0.00)}                      \\ \hline
\multirow{2}{*}{\textbf{Religion}}        & adjectives vs religion terms                                & 1.05 (0.00)                      & \textbf{1.28 (0.01)}                      & 1.04 (0.00)                      & \textbf{1.20 (0.00)}                      \\
                                          & adjectives vs lastnames                                     & 0.93 (0.00)                      & \textbf{1.55 (0.00)}                      & 0.95 (0.00)                      & \textbf{1.41 (0.00)}                     \\ \hline

\textbf{Occupation}                            & adjectives vs urban, rural occupations                                         & -0.08 (0.59)                      & \textbf{1.58 (0.00)}                     & -0.13 (0.88)                      & \textbf{1.42 (0.00)}                      \\ \hline
\end{tabular}
\caption{WEAT and SEAT bias measurements (with p-values in parentheses) for tests with translated versus language-specific word lists. Highlighted values point towards the observation that more bias is captured in language-specific curated word lists. 
}
\label{tab:trans-vs-crafted}
\end{table*}

For the WEAT Hindi test set, 
we create another test to quantify gender bias across neutral adjectives, based on societal biases in Indic society \cite{gundugurti2015psy}.  
Appendix \ref{app:wordlists} lists all word sets used in each WEAT test.
Further, in Section \ref{section:gender_in_hindi}, we see that there are four specific gendered word groups in Hindi, all of which are meaningfully gendered and important to be encoded and retained in our representations. For each such group, we construct an independent ``Meaningful Encoding (ME)'' WEAT test ~\cite{dev-etal-2021-oscar} (see word lists in Appendix \ref{app:wordlists}). A Meaningful Encoding WEAT test uses attribute lists of words having meaningful gendered information (like gendered verbs) which should be captured by representations. The importance of this is two-fold: (i) it allows us to verify if meaningful gendered information is encoded in our representations, and (ii) compare with biased associations (measured by WEAT) to gauge the overall magnitude of bias versus information about an attribute captured by our embeddings. 

In Table \ref{tab:weatandseatforall} we observe that for 300 dimensional Hindi GloVe embeddings \cite{kumar2020passage}, significant bias is observed using the three WEAT tests (\textit{Language-Specific}) for binary gender and adjectives, science v/s arts, and maths v/s arts. Each score is over 1.00, and similar valued to WEAT tests for meaningful information encoding (ME scores in Table \ref{tab:weatandseatforall}), which highlights how the magnitude of bias encoded is high. Of the four meaningful information encodings, the weakest association is seen among gendered entities, owing to how prone they are to ambiguous usage across different regions (Section \ref{section:gender_in_hindi}).

To develop SEAT tests in Hindi, similar to \cite{may-etal-2019-measuring}, we construct a list of sentence templates and fits each target word from a WEAT target list to construct SEAT target lists for each part of the speech category. We used the Hindi translations of the semantically neutral templates provided in \cite{may-etal-2019-measuring}. However, we remove some ambiguous translations and we add other templates based on colloquial usage (Appendix \ref{app: seat}) and word lists created for WEAT (Appendix \ref{app:wordlists}). 
We conduct SEAT tests upon 300 dimensional Hindi GloVe embeddings and Hindi ELMo \cite{kumar2020passage}. 

Table \ref{tab:weatandseatforall} demonstrates that for GloVe, the SEAT scores report significant bias for all tests, while for 
ELMo, the bias is mainly measured in tests with binary gender and adjectives.
From Table \ref{tab:trans-vs-crafted} we note that the \textit{Language-Specific} word lists record higher amounts of bias than the translated test.
This highlights the significance of constructing the word lists in WEAT tests while keeping language and cultural considerations in mind.



\begin{table*}[]
\small
\centering
\begin{tabular}{|cc|c|c|cc|}
\hline
\multicolumn{2}{|c|}{\multirow{2}{*}{\textbf{Attribute}}}                               & \multirow{2}{*}{\textbf{Description}}        & \textbf{WEAT}        & \multicolumn{2}{c|}{\textbf{SEAT}}               \\ \cline{4-6} 
\multicolumn{2}{|c|}{}                                                                  &                                              & \textbf{GloVe}       & \textbf{GloVe}                   & \textbf{ELMo} \\ \hline
\multicolumn{1}{|c|}{\multirow{7}{*}{\textbf{Gender}}}   & \multirow{3}{*}{BM}          & maths, arts vs male, female                  & 1.12 (0.01)          & 1.14 (0.00)                      & 0.17          \\
\multicolumn{1}{|c|}{}                                   &                              & science, arts vs male, female                & 1.13 (0.01)          & 1.03 (0.00)                      & 0.14          \\
\multicolumn{1}{|c|}{}                                   &                              & adjectives vs male, female                   & 1.21 (0.02)          & 1.19 (0.00)                      & 1.37          \\ \cline{2-6} 
\multicolumn{1}{|c|}{}                                   & \multirow{4}{*}{\textit{ME}} & \textit{gendered verbs vs male, female}      & \textit{1.87 (0.00)} & \textit{1.84 (0.00)}             & \textit{1.66} \\
\multicolumn{1}{|c|}{}                                   &                              & \textit{gendered adjectives vs male, female} & \textit{1.70 (0.00)} & \textit{1.63 (0.00)}             & \textit{1.78} \\
\multicolumn{1}{|c|}{}                                   &                              & \textit{gendered entities vs male, female}   & \textit{1.14 (0.01)} & \textit{1.12 (0.00)}             & \textit{1.77} \\
\multicolumn{1}{|c|}{}                                   &                              & \textit{gendered titles vs male, female}     & \textit{1.92 (0.00)} & \textit{1.86 (0.00)}             & \textit{1.64} \\ \hline
\multicolumn{1}{|c|}{\multirow{2}{*}{\textbf{Caste}}}    & \multirow{2}{*}{BM}          & occupations vs caste                         & 1.44 (0.00)          & 1.26 (0.00)                      & 0.89          \\
\multicolumn{1}{|c|}{}                                   &                              & adjectives vs caste                          & 1.52 (0.00)          & 1.40 (0.00)                      & 0.48          \\ \hline
\multicolumn{1}{|c|}{\multirow{3}{*}{\textbf{Religion}}} & \multirow{2}{*}{BM}          & adjectives vs religion terms                 & 1.28 (0.01)          & 1.20 (0.00)                      & 0.75          \\
\multicolumn{1}{|c|}{}                                   &                              & adjectives vs lastnames                      & 1.55 (0.00)          & 1.41 (0.00)                      & 1.02          \\ \cline{2-6} 
\multicolumn{1}{|c|}{}                                   & \textit{ME}                  & \textit{religious entities vs religion}      & \textit{1.75 (0.00)} & \textit{1.69 (0.00)}             & \textit{1.23} \\ \hline
\multicolumn{1}{|c|}{\textbf{Occupation}}                & BM                           & adjectives vs urban, rural occupations       & 1.58 (0.00)          & \multicolumn{1}{c}{1.42 (0.00)} & 1.12          \\ \hline
\end{tabular}
\caption{Bias measurements (with p-values in parentheses) for gender, caste, religion and occupation bias. 
These results are for \textit{Language-Specific} word lists; BM: Bias Measuring test, ME: Meaningful Encoding test.}
\label{tab:weatandseatforall}
\vspace{-0.5cm}
\end{table*}

\vspace{-0.25cm}
\subsection{Caste Bias}
To evaluate Hindi representations for caste bias, 
we build two WEAT tests and two corresponding SEAT tests using the last names that are statistically more associated with stereotypically lower and upper castes. For lower castes, we randomly sample lower caste names from the list of scheduled castes provided by the Department of Social Justice and Empowerment in India (Appendix \ref{app:wordlists}). Our first test is based upon detecting biased association of occupations with `upper' castes (the upper strata of castes) and `lower' castes. Note that, some caste names mean certain occupations themselves in Hindi language. For example `kumhar' means both a lower caste and the occupation of pottery. We ensure that target word lists have no ambiguous entities. 
Another WEAT test we build is based upon positive and negative adjectives association with caste names. For the \textit{Translated} version, we take Hindi translations of words from \citet{caliskan2017semantics} for detecting racial bias. For \textit{Language-specific} test we curate a new word list of adjectives (Appendix \ref{app:wordlists}) based on words used popularly as positive or negative in Hindi \cite{thorat2007caste}.

Table \ref{tab:weatandseatforall} highlights that there is significant caste related perception bias. For both WEAT and SEAT tests, the measured biases are over 1.2 for GloVe embeddings. 
The results in Table \ref{tab:trans-vs-crafted} compare the \textit{Translated} and \textit{Language-Specific} adjective word lists. For WEAT test, the bias measured by \textit{Translated} word lists are less than half of that measured by \textit{Language-Specific} word lists, emphasizing the importance of creating socially-aware \textit{Language-Specific} word lists which correlate better with the society and culture the language is associated with. 

\subsection{Religion Bias}

We construct two WEAT and two SEAT tests to detect religion associated biases in Hindi embeddings. Our first test is based upon associating positive and negative adjectives to religious entities. One attribute list consists of Hindu religious entities and one consists of Muslim religious entities. In our second test, we associate adjectives with lastnames. This stems from the distinct last names commonly used by the two populations (see Appendix \ref{app:wordlists}). Similar to Caste bias detection, we experiment with the \textit{Translated} and \textit{Language-Specific} adjective lists.  
Further, we evaluate if religious information which is correctly associated is learnt by the representations (for example, mosque being the place of worship in Islam should be associated with it in representations). 
For this, we create a meaningful encoding (ME) test for religious information (see word lists in Appendix).

In Table \ref{tab:weatandseatforall} we see using the WEAT and SEAT scores for 300-dimensional GloVe embeddings that significant religious bias with respect to the positive and negative perception is detected. 
Table \ref{tab:trans-vs-crafted} compares the measured bias in case of \textit{Translated} and \textit{Language-Specific} adjective word lists, with the latter capturing 
significantly larger bias.
\subsection{Rural v/s Urban Occupation Bias}
Besides,  we detect bias in urban and rural occupations, which is prevalent in Indic society - with urban occupations seen as better, richer, more desirable, and even of a higher social status. We construct WEAT and SEAT tests where the attribute list consists of lists of urban occupations and rural occupations and the target lists consisted of polarized adjectives 
(Appendix \ref{app:wordlists}). 

Table \ref{tab:weatandseatforall} illustrates with WEAT and SEAT scores the biased associations of perception between urban and rural occupations. For both GloVe and ELMo embeddings, we observe significant ($>$ 1.0) bias with the WEAT test, 
highlighting the presence of occupation associated bias. 




\section{Conclusion}
\vspace{-0.3cm}
Biases are inherently complex as are their encodings in language representations. Their detection needs to take into account a multitude of factors - the language and its grammar, the regions it is spoken in, as well as the history and culture of the region. We demonstrate here how a predetermined set of biases and a peripheral evaluation consisting of a narrow perspective of how biases manifest themselves is not sufficient to achieve fair representations across the globe. 

Our work is limited by the scarcity of robust language models in Hindi language, as well as dedicated word lists for different language tasks in Hindi language. Hence, a number of extrinsic tests and experiments for bias evaluation could not be performed to evaluate bias more extensively. We thus focus here only on intrinsic measurements of bias which may not be correlated with bias expressed in downstream tasks~\cite{goldfarb-tarrant-etal-2019-plan,cao2022intrinsic}. We leave investigations regarding the same to future work.
Furthermore, we acknowledge that our analysis of gender associated biases is limited to binary gender and our intrinsic evaluations require discrete categorizations~\cite{dev-etal-2021-harms,antoniak-mimno-2021-bad}.
Finally, despite the limitations, we believe our work lays down some fundamentals with respect to evaluating biases across languages and associated cultures.

\section*{Broader Impact}
Language models, with their widespread applications impact people across the world. This makes it imperative that associated harms be understood not just for the Western world and with a focus on English language models, but also across languages and cultures. With this work, we highlight the importance of social and cultural awareness for the same. Bias detection methods need this cultural expertise and can then be followed by adapted mitigation methods (some possible adapted methods discussed in Appendix \ref{app:debiasingwordlists}). With this work, we demonstrate how translations of words is not sufficient for capturing biases across languages, and thus highlight the need for development of strategies with specific languages and cultures in mind.

\section*{Acknowledgements}
This work was supported by 
NSF IIS-1927554 and NSF grant
\#2030859 to the Computing Research Association for the CIFellows Project. 
We thank the anonymous reviewers, and members of UCLA-NLP and
Plus labs for their feedback.

\bibliographystyle{acl_natbib}
\bibliography{acl2021, anthology_new}

\begin{thebibliography}{39}
\expandafter\ifx\csname natexlab\endcsname\relax\def\natexlab#1{#1}\fi

\bibitem[{Amutha(2017)}]{amutha2017roots}
D.~Amutha. 2017.
\newblock \href {https://ssrn.com/abstract=2906950} {The roots of gender
  inequality in india}.

\bibitem[{Antoniak and Mimno(2021)}]{antoniak-mimno-2021-bad}
Maria Antoniak and David Mimno. 2021.
\newblock \href {https://doi.org/10.18653/v1/2021.acl-long.148} {Bad seeds:
  Evaluating lexical methods for bias measurement}.
\newblock In \emph{Proceedings of the 59th Annual Meeting of the Association
  for Computational Linguistics and the 11th International Joint Conference on
  Natural Language Processing (Volume 1: Long Papers)}, pages 1889--1904,
  Online. Association for Computational Linguistics.

\bibitem[{Banerjee and Knight(1985)}]{banerjee1985caste}
Biswajit Banerjee and John~B Knight. 1985.
\newblock Caste discrimination in the indian urban labour market.
\newblock \emph{Journal of development Economics}, 17(3):277--307.

\bibitem[{Bhagwat et~al.(2019)Bhagwat, Nagarkar, Paramane, and
  Jindam}]{Bhagwat2019ReviewOC}
Varad Bhagwat, Mrunali~N Nagarkar, Pooja Paramane, and Shrikant Jindam. 2019.
\newblock Review of chatbot system in hindi language.
\newblock In \emph{Review of Chatbot System in Hindi Language}.

\bibitem[{Bojanowski et~al.(2017)Bojanowski, Grave, Joulin, and
  Mikolov}]{bojanowski2017enriching}
Piotr Bojanowski, Edouard Grave, Armand Joulin, and Tomas Mikolov. 2017.
\newblock Enriching word vectors with subword information.
\newblock \emph{Transactions of the Association for Computational Linguistics},
  5:135--146.

\bibitem[{Bolukbasi et~al.(2016)Bolukbasi, Chang, Zou, Saligrama, and
  Kalai}]{NIPS2016_a486cd07}
Tolga Bolukbasi, Kai-Wei Chang, James~Y Zou, Venkatesh Saligrama, and Adam~T
  Kalai. 2016.
\newblock \href
  {https://proceedings.neurips.cc/paper/2016/file/a486cd07e4ac3d270571622f4f316ec5-Paper.pdf}
  {Man is to computer programmer as woman is to homemaker? debiasing word
  embeddings}.
\newblock In \emph{Advances in Neural Information Processing Systems},
  volume~29. Curran Associates, Inc.

\bibitem[{Caliskan et~al.(2017)Caliskan, Bryson, and
  Narayanan}]{caliskan2017semantics}
Aylin Caliskan, Joanna~J Bryson, and Arvind Narayanan. 2017.
\newblock Semantics derived automatically from language corpora contain
  human-like biases.
\newblock \emph{Science}, 356(6334):183--186.

\bibitem[{Cao et~al.(2022)Cao, Pruksachatkun, Chang, Gupta, Kumar, Dhamala, and
  Galstyan}]{cao2022intrinsic}
Yang~Trista Cao, Yada Pruksachatkun, Kai-Wei Chang, Rahul Gupta, Varun Kumar,
  Jwala Dhamala, and Aram Galstyan. 2022.
\newblock \href {https://doi.org/10.48550/ARXIV.2203.13928} {On the intrinsic
  and extrinsic fairness evaluation metrics for contextualized language
  representations}.

\bibitem[{Ch{\'a}vez~Mulsa and
  Spanakis(2020)}]{chavez-mulsa-spanakis-2020-evaluating}
Rodrigo~Alejandro Ch{\'a}vez~Mulsa and Gerasimos Spanakis. 2020.
\newblock \href {https://aclanthology.org/2020.gebnlp-1.6} {Evaluating bias in
  {D}utch word embeddings}.
\newblock In \emph{Proceedings of the Second Workshop on Gender Bias in Natural
  Language Processing}, pages 56--71, Barcelona, Spain (Online). Association
  for Computational Linguistics.

\bibitem[{Cheung and Chan(2007)}]{doi:10.1177/1028315306291538}
Hoi~Yan Cheung and Alex W.~H. Chan. 2007.
\newblock \href {https://doi.org/10.1177/1028315306291538} {How culture affects
  female inequality across countries: An empirical study}.
\newblock \emph{Journal of Studies in International Education}, 11(2):157--179.

\bibitem[{Deshpande(2010)}]{deshpande2010history}
Manali~S Deshpande. 2010.
\newblock History of the indian caste system and its impact on india today.
\newblock \emph{CalPoly Student Research}.

\bibitem[{Dev et~al.(2021{\natexlab{a}})Dev, Li, Phillips, and
  Srikumar}]{dev-etal-2021-oscar}
Sunipa Dev, Tao Li, Jeff~M Phillips, and Vivek Srikumar. 2021{\natexlab{a}}.
\newblock \href {https://doi.org/10.18653/v1/2021.emnlp-main.411} {{OSC}a{R}:
  Orthogonal subspace correction and rectification of biases in word
  embeddings}.
\newblock In \emph{Proceedings of the 2021 Conference on Empirical Methods in
  Natural Language Processing}, pages 5034--5050, Online and Punta Cana,
  Dominican Republic. Association for Computational Linguistics.

\bibitem[{Dev et~al.(2021{\natexlab{b}})Dev, Monajatipoor, Ovalle, Subramonian,
  Phillips, and Chang}]{dev-etal-2021-harms}
Sunipa Dev, Masoud Monajatipoor, Anaelia Ovalle, Arjun Subramonian, Jeff
  Phillips, and Kai-Wei Chang. 2021{\natexlab{b}}.
\newblock \href {https://doi.org/10.18653/v1/2021.emnlp-main.150} {Harms of
  gender exclusivity and challenges in non-binary representation in language
  technologies}.
\newblock In \emph{Proceedings of the 2021 Conference on Empirical Methods in
  Natural Language Processing}, pages 1968--1994, Online and Punta Cana,
  Dominican Republic. Association for Computational Linguistics.

\bibitem[{Dev and Phillips(2019)}]{dev2019attenuating}
Sunipa Dev and Jeff Phillips. 2019.
\newblock Attenuating bias in word vectors.
\newblock In \emph{The 22nd International Conference on Artificial Intelligence
  and Statistics}, pages 879--887. PMLR.

\bibitem[{Devlin et~al.(2019)Devlin, Chang, Lee, and
  Toutanova}]{devlin-etal-2019-bert}
Jacob Devlin, Ming-Wei Chang, Kenton Lee, and Kristina Toutanova. 2019.
\newblock \href {https://doi.org/10.18653/v1/N19-1423} {{BERT}: Pre-training of
  deep bidirectional transformers for language understanding}.
\newblock In \emph{Proceedings of the 2019 Conference of the North {A}merican
  Chapter of the Association for Computational Linguistics: Human Language
  Technologies, Volume 1 (Long and Short Papers)}, pages 4171--4186,
  Minneapolis, Minnesota. Association for Computational Linguistics.

\bibitem[{Ghosh et~al.(2021)Ghosh, Baker, Jurgens, and
  Prabhakaran}]{ghosh2021detecting}
Sayan Ghosh, Dylan Baker, David Jurgens, and Vinodkumar Prabhakaran. 2021.
\newblock \href {http://arxiv.org/abs/2104.06999} {Detecting cross-geographic
  biases in toxicity modeling on social media}.

\bibitem[{Goldfarb-Tarrant et~al.(2019)Goldfarb-Tarrant, Feng, and
  Peng}]{goldfarb-tarrant-etal-2019-plan}
Seraphina Goldfarb-Tarrant, Haining Feng, and Nanyun Peng. 2019.
\newblock \href {https://doi.org/10.18653/v1/N19-4016} {Plan, write, and
  revise: an interactive system for open-domain story generation}.
\newblock In \emph{Proceedings of the 2019 Conference of the North {A}merican
  Chapter of the Association for Computational Linguistics (Demonstrations)},
  pages 89--97, Minneapolis, Minnesota. Association for Computational
  Linguistics.

\bibitem[{Gundugurti et~al.(2015)Gundugurti, Vidya, and
  Sriramya}]{gundugurti2015psy}
Prasad Gundugurti, KL~Vidya, and V~Sriramya. 2015.
\newblock \href {https://doi.org/10.4103/0019-5545.161480} {The indian "girl"
  psychology: A perspective}.
\newblock \emph{Indian journal of psychiatry}, 57:S212--5.

\bibitem[{Gupta(2021)}]{GUPTA2021}
Khushi Gupta. 2021.
\newblock \href {http://www.inquiriesjournal.com/a?id=1868} {Stereotypes in
  bollywood cinema: Does article 15 reinforce the dalit narrative?}
\newblock In \emph{Inquiries Journal [Online]}.

\bibitem[{Hall(2002)}]{inbook}
Kira Hall. 2002.
\newblock \href {https://doi.org/10.13140/2.1.4155.8086} {\emph{"Unnatural"
  Gender in Hindi}}, pages 133--162. Oxford University Press.

\bibitem[{Khadilkar et~al.(2021)Khadilkar, KhudaBukhsh, and
  Mitchell}]{KHADILKAR2021100409}
Kunal Khadilkar, Ashiqur~R. KhudaBukhsh, and Tom~M. Mitchell. 2021.
\newblock \href {https://doi.org/https://doi.org/10.1016/j.patter.2021.100409}
  {Gender bias, social bias, and representation: 70 years of bhollywood}.
\newblock \emph{Patterns}, page 100409.

\bibitem[{Kumar et~al.(2020)Kumar, Kumar, Kanojia, and
  Bhattacharyya}]{kumar2020passage}
Saurav Kumar, Saunack Kumar, Diptesh Kanojia, and Pushpak Bhattacharyya. 2020.
\newblock “a passage to india”: Pre-trained word embeddings for indian
  languages.
\newblock In \emph{Proceedings of the 1st Joint Workshop on Spoken Language
  Technologies for Under-resourced languages (SLTU) and Collaboration and
  Computing for Under-Resourced Languages (CCURL)}, pages 352--357.

\bibitem[{Kumar et~al.(2019)Kumar, Mahata, Aggarwal, Chugh, Maheshwari, and
  Shah}]{kumar2019bhaav}
Yaman Kumar, Debanjan Mahata, Sagar Aggarwal, Anmol Chugh, Rajat Maheshwari,
  and Rajiv~Ratn Shah. 2019.
\newblock Bhaav-a text corpus for emotion analysis from hindi stories.
\newblock \emph{arXiv preprint arXiv:1910.04073}.

\bibitem[{Kunchukuttan et~al.(2017)Kunchukuttan, Mehta, and
  Bhattacharyya}]{kunchukuttan2017iit}
Anoop Kunchukuttan, Pratik Mehta, and Pushpak Bhattacharyya. 2017.
\newblock The {IIT Bombay English-Hindi} parallel corpus.
\newblock \emph{arXiv preprint arXiv:1710.02855}.

\bibitem[{Kurpicz-Briki(2020)}]{kurpicz2020cultural}
Mascha Kurpicz-Briki. 2020.
\newblock Cultural differences in bias? origin and gender bias in pre-trained
  german and french word embeddings.
\newblock \emph{Proceedings of the 5th Swiss Text Analytics Conference
  (SwissText) \& 16th Conference on Natural Language Processing (KONVENS)}.

\bibitem[{Lauscher et~al.(2020)Lauscher, Takieddin, Ponzetto, and
  Glava{\v{s}}}]{lauscher-etal-2020-araweat}
Anne Lauscher, Rafik Takieddin, Simone~Paolo Ponzetto, and Goran Glava{\v{s}}.
  2020.
\newblock \href {https://aclanthology.org/2020.wanlp-1.17} {{A}ra{WEAT}:
  Multidimensional analysis of biases in {A}rabic word embeddings}.
\newblock In \emph{Proceedings of the Fifth Arabic Natural Language Processing
  Workshop}, pages 192--199, Barcelona, Spain (Online). Association for
  Computational Linguistics.

\bibitem[{Liang et~al.(2020)Liang, Dufter, and
  Sch{\"u}tze}]{liang-etal-2020-monolingual}
Sheng Liang, Philipp Dufter, and Hinrich Sch{\"u}tze. 2020.
\newblock \href {https://doi.org/10.18653/v1/2020.coling-main.446} {Monolingual
  and multilingual reduction of gender bias in contextualized representations}.
\newblock In \emph{Proceedings of the 28th International Conference on
  Computational Linguistics}, pages 5082--5093, Barcelona, Spain (Online).
  International Committee on Computational Linguistics.

\bibitem[{Madaan et~al.(2018)Madaan, Mehta, Agrawaal, Malhotra, Aggarwal,
  Gupta, and Saxena}]{Madaan2018AnalyzeDA}
Nishtha Madaan, Sameep Mehta, Taneea~S. Agrawaal, Vrinda Malhotra, Aditi
  Aggarwal, Yatin Gupta, and Mayank Saxena. 2018.
\newblock Analyze, detect and remove gender stereotyping from bollywood movies.
\newblock In \emph{FAccT}.

\bibitem[{Manzini et~al.(2019)Manzini, Yao~Chong, Black, and
  Tsvetkov}]{manzini-etal-2019-black}
Thomas Manzini, Lim Yao~Chong, Alan~W Black, and Yulia Tsvetkov. 2019.
\newblock \href {https://doi.org/10.18653/v1/N19-1062} {Black is to criminal as
  caucasian is to police: Detecting and removing multiclass bias in word
  embeddings}.
\newblock In \emph{Proceedings of the 2019 Conference of the North {A}merican
  Chapter of the Association for Computational Linguistics: Human Language
  Technologies, Volume 1 (Long and Short Papers)}, pages 615--621, Minneapolis,
  Minnesota. Association for Computational Linguistics.

\bibitem[{May et~al.(2019)May, Wang, Bordia, Bowman, and
  Rudinger}]{may-etal-2019-measuring}
Chandler May, Alex Wang, Shikha Bordia, Samuel~R. Bowman, and Rachel Rudinger.
  2019.
\newblock \href {https://doi.org/10.18653/v1/N19-1063} {On measuring social
  biases in sentence encoders}.
\newblock In \emph{Proceedings of the 2019 Conference of the North {A}merican
  Chapter of the Association for Computational Linguistics: Human Language
  Technologies, Volume 1 (Long and Short Papers)}, pages 622--628, Minneapolis,
  Minnesota. Association for Computational Linguistics.

\bibitem[{Nangia et~al.(2020)Nangia, Vania, Bhalerao, and
  Bowman}]{nangia-etal-2020-crows}
Nikita Nangia, Clara Vania, Rasika Bhalerao, and Samuel~R. Bowman. 2020.
\newblock \href {https://doi.org/10.18653/v1/2020.emnlp-main.154}
  {{C}row{S}-pairs: A challenge dataset for measuring social biases in masked
  language models}.
\newblock In \emph{Proceedings of the 2020 Conference on Empirical Methods in
  Natural Language Processing (EMNLP)}, pages 1953--1967, Online. Association
  for Computational Linguistics.

\bibitem[{Pennington et~al.(2014)Pennington, Socher, and
  Manning}]{pennington2014glove}
Jeffrey Pennington, Richard Socher, and Christopher~D Manning. 2014.
\newblock Glove: Global vectors for word representation.
\newblock In \emph{Proceedings of the 2014 conference on empirical methods in
  natural language processing (EMNLP)}, pages 1532--1543.

\bibitem[{Pujari et~al.(2019)Pujari, Mittal, Padhi, Jain, Jadon, and
  Kumar}]{10.1145/3377713.3377792}
Arun~K. Pujari, Ansh Mittal, Anshuman Padhi, Anshul Jain, Mukesh Jadon, and
  Vikas Kumar. 2019.
\newblock \href {https://doi.org/10.1145/3377713.3377792} {Debiasing gender
  biased hindi words with word-embedding}.
\newblock In \emph{Proceedings of the 2019 2nd International Conference on
  Algorithms, Computing and Artificial Intelligence}, ACAI 2019, page
  450–456, New York, NY, USA. Association for Computing Machinery.

\bibitem[{Pundir and Singh(2019)}]{articlepundir}
Ishita Pundir and Alankrita Singh. 2019.
\newblock Portrayal of women in indian fiction.
\newblock Volume 09:137--141.

\bibitem[{Rao and Devi(2018)}]{rao2018eventxtract}
Pattabhi~RK Rao and Sobha~Lalitha Devi. 2018.
\newblock Eventxtract-il: Event extraction from newswires and social media text
  in indian languages@ fire 2018-an overview.
\newblock In \emph{FIRE (Working Notes)}, pages 282--290.

\bibitem[{Sambasivan et~al.(2020)Sambasivan, Arnesen, Hutchinson, and
  Prabhakaran}]{Sambasivan2020NonportabilityOA}
Nithya Sambasivan, Erin Arnesen, Ben Hutchinson, and Vinodkumar Prabhakaran.
  2020.
\newblock Non-portability of algorithmic fairness in india.
\newblock \emph{ArXiv}, abs/2012.03659.

\bibitem[{Thorat and Newman(2007)}]{thorat2007caste}
Sukhadeo Thorat and Katherine~S Newman. 2007.
\newblock Caste and economic discrimination: causes, consequences and remedies.
\newblock \emph{Economic and Political Weekly}, pages 4121--4124.

\bibitem[{Zhao et~al.(2020)Zhao, Mukherjee, Hosseini, Chang, and
  Hassan~Awadallah}]{zhao-etal-2020-gender}
Jieyu Zhao, Subhabrata Mukherjee, Saghar Hosseini, Kai-Wei Chang, and Ahmed
  Hassan~Awadallah. 2020.
\newblock \href {https://doi.org/10.18653/v1/2020.acl-main.260} {Gender bias in
  multilingual embeddings and cross-lingual transfer}.
\newblock In \emph{Proceedings of the 58th Annual Meeting of the Association
  for Computational Linguistics}, pages 2896--2907, Online. Association for
  Computational Linguistics.

\bibitem[{Zhou et~al.(2019)Zhou, Shi, Zhao, Huang, Chen, Cotterell, and
  Chang}]{zhou-etal-2019-examining}
Pei Zhou, Weijia Shi, Jieyu Zhao, Kuan-Hao Huang, Muhao Chen, Ryan Cotterell,
  and Kai-Wei Chang. 2019.
\newblock \href {https://doi.org/10.18653/v1/D19-1531} {Examining gender bias
  in languages with grammatical gender}.
\newblock In \emph{Proceedings of the 2019 Conference on Empirical Methods in
  Natural Language Processing and the 9th International Joint Conference on
  Natural Language Processing (EMNLP-IJCNLP)}, pages 5276--5284, Hong Kong,
  China. Association for Computational Linguistics.

\end{thebibliography}

\newpage
\appendix
\section*{{\Large\selectfont{Appendix}}}
\label{appendix}

\section{Limitations}
\label{app: limitations}
As acknowledged in the paper, our work is severely limited by the scarcity of language models, dedicated word lists, and language tasks for Hindi language. A number of tests and experiments could not be performed to evaluate bias more extensively. Furthermore, the lack of established language tasks and datasets in Hindi made it difficult to analyze the extrinsic bias in downstream tasks. Although this limits our evaluations of bias in this work, with more work like this and development of more language tools for Hindi, this can be overcome. 
We further emphasize that while we have evaluated some biases, these are not the only biases present in the Indian society or Hindi language. We merely provide evaluations for some that are strongly present in the literature and thus in the language representations as well. 

Since this work highlights various biases and the words commonly associated with it, it can potentially be triggering to persons. However, it is important to study these biases and their impact on language tools in order to mitigate their effect. 

\section{WEAT}
\label{app: weat}
Let $X$ and $Y$ be equal-sized sets of target concept embeddings and let $A$ and $B$ be sets of
attribute embeddings. Let $cos(a,b)$ denote the cosine similarity between vectors $a$ and $b$. The test statistic is a difference between sums
over the respective target concepts,
\begin{align}
s(X,Y,A,B) = \sum_{x\in X} s(x,A,B) - \sum_{y\in Y} s(y,A,B)
\end{align}
where the quantity, $s(w,A,B)$ is calculated using cosine similarity as follows:
\begin{equation}
s(w,A,B) = \frac{\sum_{a\in A}cos(w,a)}{|A|} - \frac{\sum_{b\in B}cos(w,b)}{|B|}
\end{equation}
The amount of bias in WEAT is analyzed by effect size $d$ calculated as:
\begin{equation}
d = \frac{mean_{x\in X}s(x,A,B) - mean_{y\in Y}s(y,A,B)}{stddev_{w\in X\cup Y}s(w,A,B)}
\end{equation}
In order to compute the test significance (p-value), $X$ and $Y$ lists are merged together, and 10,000 permutations, $P$ of the combined list is generated. For the $i$-th list in $P$,  it is split in new pairs of $X_{i}$ and $Y_{i}$ lists. Then the test statistic equation is used to calculate the p-value in the following way:

\begin{equation}
p = \frac{\sum_{i\in P}[s(X_{i}, Y_{i}, A, B)] > s(X, Y, A, B)]}{|P|}
\end{equation}

\section{WEAT word lists}
\label{app:wordlists}
In our bias detection methods using WEAT, we constructed WEAT tests for Gender, Caste, Religion and Occupation (Rural v/s Urban occupations) biases. Since the direct translations of word lists from \cite{caliskan2017semantics} did not provide us with any significant evidence of bias, we constructed the word lists ourselves based upon popular Hindi words usage.
Refer to Table \ref{tab:genderwordlists} for the word lists used to detect gender bias. We obtain the \href{https://www.proz.com/personal-glossaries/97964-english-to-hindi-physics?page=2}{science}, \href{https://hindimiddleeast.com/mathematics-vocabulary-in-english-and-hindi/}{maths}, \href{https://www.hindipod101.com/blog/2020/05/17/guide-to-hindi-grammatical-gender/}{gendered words}, \href{https://www.hindipod101.com/blog/2020/03/24/top-100-hindi-adjectives/}{gendered adjectives} and \href{https://www.hindipod101.com/blog/2020/03/24/top-100-hindi-adjectives/}{occupations} from online Hindi resources. We refer to Wikipedia for glossary of \href{https://en.wikipedia.org/wiki/Glossary_of_Hinduism_terms}{Hinduism} entities and \href{ https://en.wikipedia.org/wiki/Glossary_of_Islam}{Islamic} entities. For the list of castes, we refer to the list of scheduled \href{http://socialjustice.nic.in/UserView/index?mid=76750}{castes} provided by the Department of Social Justice and Empowerment in India. For the lastnames we refer to the popular \href{https://www.familyeducation.com/baby-names/browse-origin/surname/muslim?page=3}{Islamic} lastnames and \href{https://www.momjunction.com/articles/popular-indian-last-names-for-your-baby_00334734/}{Hindu} lastnames provided by online resources.

In addition to the bias measurement (BM), tests, we also provide meaningful encoding. (ME) tests used to capture meaningful gendered information in Hindi.

Table \ref{tab:castewordlists}, Table \ref{tab:religionwordlists} and Table \ref{tab:rural-urban} provides the word lists used in the measurement of caste, religion and occupation biases respectively.

\section{SEAT}
\label{app: seat}
We define a list of semantically neutral sentence templates for each part of speech type of words as follows:

\begin{itemize}
    \item Hindi-SEAT-name: `yeha \_ hai', `veha \_ hai', `vahan \_ hai', `yahan \_ hai', `\_ yahan hai', `\_ vahan hai', `iska naam \_ hai', `uska naam \_ h'
    \item Hindi-SEAT-common-nouns: `yeha \_ hai', `veha \_ hai', `vahan \_ hai', `yahan \_ hai', `\_ yahan hai', `\_ vahan hai', `vo \_ hai', `ye \_ hai'
    \item Hindi-SEAT-verbs: `yeha \_ hai', `veha \_ hai', `vo \_ hai', `ye \_ hai', `vahan \_ hai', `yahan \_ hai'
    \item Hindi-SEAT-adjectives: `yeha \_ hai', `veha \_ hai', `vo \_ hai', `ye \_ hai'
\end{itemize}
In other words, if the target word is adjective, we use Hindi-SEAT-adjective list of semantically bleached sentences with each WEAT target word.

\section{Debiasing}
\label{app:debiasingwordlists}
There have been notable advances towards debiasing embeddings along the direction of gender bias. Both \citet{NIPS2016_a486cd07} and \citet{dev2019attenuating} propose using linear projection to debias word embeddings, but the former in addition also equalizes word pairs about the attribute (e.g., gender) axis. 

Although we tried and adapted several existing methods for debiasing, we could not evaluate the performance of the debiasing methods on the extrinsic tasks. This is because of the scarcity of reliable Hindi language datasets, which made any form of notable inferences harder. In addition, the deep learning models were already underperforming on these Hindi datasets.

In this work we use the more general approach of linear projection as it can be adapted to several biases apart from gender. 

In the method of linear projection, all words $w\in W$ are debiased to $w'$ by being projected orthogonally to the identified bias vector $v_{B}$.

\begin{equation}
w' = w - \langle w, v_{B} \rangle v_{B}
\end{equation}

In case of hard debiasing, we required list of equalizing pairs and list of words to not debias in Hindi. However, direct translation of the word lists to Hindi did not always make sense. Since, some words like `she' and `he' had overlapping translations and both the pronouns are referred to as `veha' in Hindi. This overlapping translation is true the other way round as well, the word grandfather can be either `nana' (maternal grandfather) or `dada' (paternal grandfather). 

For languages with grammatical gender, \citet{zhou-etal-2019-examining} proposed to determine semantic gender direction. To obtain the semantic gender direction ($d_{s}$), the grammatical gender component ($d_{g}$) in the computed gender direction (obtained from PCA over gendered word pairs, $d_{PCA}$) to
make the semantic gender direction orthogonal to grammatical gender.

\begin{equation}
    d_{s} = d_{PCA} - \langle d_{PCA}, d_{g}\rangle d_{g}
\end{equation}

We use this orthogonalized gender direction to perform linear debiasing. We refer to this method as LPSG (Linear Projection with Semantic Gender).

\subsection{Debiasing Binary Gender}
The first step in debiasing using linear projection is to identify a bias subspace/vector. We experiment with different settings to identify the gender vector in Hindi, including (i) a single gender specific word pair direction of \{$\vec{naari}-\vec{nar}$\}, (ii) PCA over a list of paired gender specific words (in the form $\{\vec{m_{i}}-\vec{f_{i}}\}$). For more results with other gender directions, refer to the Appendix. Also, the word lists used in the experiment are provided in Appendix \ref{app:debiasingwordlists}.

For hard debiasing, we considered two types of gender definition word lists. In one list we included only the gender definitional pairs translated to Hindi from the original English lists (after some modifications to remove ambiguous translations). In another experiment, we added pairs of gendered verbs to the list as well.

Hindi is a language having grammatical gender. As introduced in Section \ref{section:gender_in_hindi}, we have 4 special gender directions along which we want to preserve the information. The direction for gendered verbs ($d_{v}$), adjectives ($d_{a}$), titles ($d_{t}$) and entities ($d_{e}$) were calculated by conducting PCA over the word lists (App \ref{app:debiasingwordlists}). For LPSG method, we provide results of orthogonalizing the semantic gender with respect to verbs and adjectives directions. In another experiment, we orthogonalize the semantic gender with respect to all the 4 directions.

Table \ref{tab:genderdebiasingall} demonstrates how different gender subspaces affect the WEAT effect sizes in both bias measuring and information retention tests. Note that the single direction of \{$\vec{naari}-\vec{nar}$\} was able to debias the best upon the math and arts test. Pairwise PCA over gendered words debiased the science and arts test quite significantly with an effect size of only 0.001 after debiasing. Hard debiasing is not able to debias the first two tests in the WEAT setting, however, it reduces the effect sizes in case of SEAT (see Table \ref{tab:genderseatall}). In both WEAT and SEAT tests for neutral adjectives vs gendered words, hard debiasing performs best against any other methods. Although hard debiasing works competitively, it comes with the downside that it does not retain the gendered information in our information retention (IR) tests. Both of the LPSG variants were able to debias competitively while at the same time were best in retaining the gendered information of IR tests.




\subsection{Debiasing Caste}
Similar to debiasing gender, for caste we first begin with determining the caste direction with a two words, one stereotypically considered upper caste and one lower: \{$\vec{ghasiya}-\vec{pandit}$\}. We also try with the direction \{$\vec{ghasiya}-\vec{desai}$\}. Since castes do not occur in pairs, a set of word pairs cannot be meaningfully constructed as done with binary gender in English~\cite{NIPS2016_a486cd07}. Hence, we compose lists of stereotypically upper and lower castes, and conduct PCA over the combined list to obtain the vector of caste bias. Refer to Appendix \ref{app:debiasingwordlists} for the word lists used in the experiment. In Table \ref{tab:castedebias} we can observe that the linear debiasing using the single direction of \{$\vec{ghasiya}-\vec{desai}$\} is unable to debias competitively when compared with the other two methods. Note that the single direction of \{$\vec{ghasiya}-\vec{pandit}$\} is able to debias better than PCA over list of caste names. 

\subsection{Debiasing Religion}
In order to mitigate religious biases in Hindi, we acknowledge how in Indian culture, the religion of a person is generally identifiable by their last names. We thus, utilize last names to determine the direction of bias. We use both (i) a single set of common last names \{$\vec{acharya}-\vec{nasir}$\}, and (ii) a set of hindu and muslim entities. 

Another religion direction is calculated by combining word lists of Hindu and Muslim lastnames and then conducting PCA over them, we call this religion bias direction as $d_{last}$. The words lists are provided in Appendix \ref{app:debiasingwordlists}.

In Hindi language, various religious entities are inherently associated with a particular religion, for example, “Bible is to Christianity as Bhagwad Gita is to Hinduism” is not bias. To accomodate for such cases, we again take motivation from \cite{zhou-etal-2019-examining} to obtain a direction $d_{ent}$ from the entities word lists (Appendix \ref{app:debiasingwordlists}) and keep the religion direction calculation calculated by Hindu and Muslim lastnames $d_{last}$, orthogonal to it.

\begin{equation}
    d_{last}' = d_{last} - \langle d_{last},d_{ent}\rangle d_{ent}
\end{equation}

We believe that if we debias words using $d_{last}'$ as bias direction, we should be able to preserve the knowledge of religion information retention test and debias competitively.

\begin{table*}
\small
\centering
\begin{tabular}{|c|c|l|ll|ll|ll|}
\hline
\multicolumn{2}{|c|}{\multirow{2}{*}{\textbf{\begin{tabular}[c]{@{}c@{}}Description\\ (vs male, female)\end{tabular}}}} & \multicolumn{1}{c|}{\multirow{2}{*}{\textbf{\begin{tabular}[c]{@{}c@{}}Original\\ WEAT\end{tabular}}}} & \multicolumn{2}{c|}{\textbf{Linear Projection}}                            & \multicolumn{2}{c|}{\textbf{Hard Debiasing}}                                                                                                                                 & \multicolumn{2}{c|}{\textbf{LPSG}}                                                                                                                                             \\ \cline{4-9} 
\multicolumn{2}{|c|}{}                                                                                                  & \multicolumn{1}{c|}{}                                                                                  & \multicolumn{1}{c}{\textbf{naari-nar}} & \multicolumn{1}{c|}{\textbf{PCA}} & \multicolumn{1}{c}{\textbf{\begin{tabular}[c]{@{}c@{}}Gen.\\ words\end{tabular}}} & \multicolumn{1}{c|}{\textbf{\begin{tabular}[c]{@{}c@{}}Gen.\\ words,verbs\end{tabular}}} & \multicolumn{1}{c}{\textbf{\begin{tabular}[c]{@{}c@{}}w/o\\ verbs \& adj\end{tabular}}} & \multicolumn{1}{c|}{\textbf{\begin{tabular}[c]{@{}c@{}}w/o\\ all dir.\end{tabular}}} \\ \hline
\multirow{3}{*}{}                                           & maths, arts                                               & 1.12 (0.01)                                                                                            & \textbf{0.44 (0.20)}                   & 0.77 (0.06)                       & 1.48 (0.00)                                                                       & 1.13 (0.00)                                                                              & 0.95 (0.03)                                                                             & 1.04 (0.02)                                                                          \\
                                                            & science, arts                                             & 1.13 (0.02)                                                                                            & 0.50 (0.18)                            & \textbf{0.00 (0.49)}              & 1.66 (0.00)                                                                       & 1.57 (0.00)                                                                              & 0.24 (0.28)                                                                             & 0.42 (0.21)                                                                          \\
                                                            & adjectives                                                & 1.22 (0.02)                                                                                            & 0.96 (0.06)                            & 0.82 (0.08)                       & \textbf{0.37 (0.27)}                                                              & 0.49 (0.20)                                                                              & 0.94 (0.05)                                                                             & 0.94 (0.049)                                                                         \\ \hline
\multirow{4}{*}{\textit{IR}}                                & \textit{gen. verbs}                                   & \textit{1.87 (0.00)}                                                                                   & \textit{\textbf{1.87 (0.00)}}          & \textit{1.79 (0.00)}              & \textit{1.12 (0.01)}                                                              & \textit{-1.18 (0.99)}                                                                    & 1.85 (0.00)                                                                             & 1.85 (0.00)                                                                          \\
                                                            & \textit{gen. adj}                                     & \textit{1.70 (0.00)}                                                                                   & \textit{1.63 (0.00)}                   & \textit{1.66 (0.00)}              & \textit{1.19 (0.00)}                                                              & \textit{0.78 (0.05)}                                                                     & \textbf{1.71 (0.00)}                                                                    & 1.75 (0.00)                                                                          \\
                                                            & \textit{gen. entities}                                & \textit{1.14 (0.01)}                                                                                   & \textit{1.01 (0.02)}                   & \textit{0.99 (0.02)}              & \textit{0.66 (0.08)}                                                              & \textit{0.27 (0.28)}                                                                     & \textbf{1.13 (0.00)}                                                                    & 1.18 (0.00)                                                                          \\
                                                            & \textit{gen. titles}                                  & \textit{1.92 (0.00)}                                                                                   & \textit{1.91 (0.00)}                   & \textit{1.89 (0.00)}              & \textit{1.19 (0.00)}                                                              & \textit{1.59 (0.00)}                                                                     & \textbf{1.92 (0.00)}                                                                    & 1.91 (0.00)                                                                          \\ \hline
\end{tabular}
\caption{Debiasing results for gender across different debiasing methods of linear projection, Bolukbasi's hard debiasing and different variants of LPSG debiasing. We provide WEAT effect sizes with p-values of the test in parentheses. PCA for Linear Projection was done on gendered word pairs. IR stands for Information Retention.}
\label{tab:genderdebiasingall}
\end{table*}

\begin{table*}
\small
\centering
\begin{tabular}{|c|c|l|ll|ll|ll|}
\hline
\multicolumn{2}{|c|}{\multirow{2}{*}{\textbf{\begin{tabular}[c]{@{}c@{}}Description\\ (vs male, female)\end{tabular}}}} & \multicolumn{1}{c|}{\multirow{2}{*}{\textbf{\begin{tabular}[c]{@{}c@{}}Original\\ SEAT\end{tabular}}}} & \multicolumn{2}{c|}{\textbf{Linear Projection}}                            & \multicolumn{2}{c|}{\textbf{Hard Debiasing}}                                                                                                                                 & \multicolumn{2}{c|}{\textbf{LPSG}}                                                                                                                                             \\ \cline{4-9} 
\multicolumn{2}{|c|}{}                                                                                                  & \multicolumn{1}{c|}{}                                                                                  & \multicolumn{1}{c}{\textbf{naari-nar}} & \multicolumn{1}{c|}{\textbf{PCA}} & \multicolumn{1}{c}{\textbf{\begin{tabular}[c]{@{}c@{}}Gen.\\ words\end{tabular}}} & \multicolumn{1}{c|}{\textbf{\begin{tabular}[c]{@{}c@{}}Gen.\\ words,verbs\end{tabular}}} & \multicolumn{1}{c}{\textbf{\begin{tabular}[c]{@{}c@{}}w/o\\ verbs \& adj\end{tabular}}} & \multicolumn{1}{c|}{\textbf{\begin{tabular}[c]{@{}c@{}}w/o\\ all dir.\end{tabular}}} \\ \hline
\multirow{3}{*}{}                                           & maths, arts                                               & 1.14 (0.00)                                                                                            & \textbf{0.64 (0.00)}                   & 0.78 (0.00)                       & 0.76 (0.00)                                                                       & 1.09 (0.00)                                                                              & 0.96 (0.00)                                                                             & 0.99 (0.00)                                                                          \\
                                                            & science, arts                                             & 1.03 (0.00)                                                                                            & 0.55 (0.00)                            & \textbf{0.07 (0.33)}              & 0.70 (0.00)                                                                       & 0.91 (0.00)                                                                              & 0.26 (0.06)                                                                             & 0.38 (0.02)                                                                          \\
                                                            & adjectives                                                & 1.19 (0.00)                                                                                            & 0.98 (0.00)                            & 0.80 (0.00)                       & \textbf{0.23 (0.30)}                                                              & 0.34 (0.31)                                                                              & 0.94 (0.00)                                                                             & 0.92 (0.00)                                                                          \\ \hline
\multirow{4}{*}{\textit{IR}}                                & \textit{gen. verbs}                                   & \textit{1.84 (0.00)}                                                                                   & \textit{\textbf{1.83 (0.00)}}          & \textit{1.67 (0.00)}              & \textit{0.31 (0.33)}                                                              & \textit{-0.70 (0.70)}                                                                    & 1.80 (0.00)                                                                             & 1.78 (0.00)                                                                          \\
                                                            & \textit{gen. adj}                                     & \textit{1.63 (0.00)}                                                                                   & \textit{1.58 (0.00)}                   & \textit{1.54 (0.00)}              & \textit{0.45 (0.17)}                                                              & \textit{0.36 (0.33)}                                                                     & \textbf{1.63 (0.00)}                                                                    & 1.67 (0.00)                                                                          \\
                                                            & \textit{gen. entities}                                & \textit{1.12 (0.00)}                                                                                   & \textit{1.02 (0.00)}                   & \textit{0.99 (0.00)}              & \textit{0.42 (0.14)}                                                              & \textit{0.45 (0.17)}                                                                     & \textbf{1.13 (0.00)}                                                                    & 1.16 (0.00)                                                                          \\
                                                            & \textit{gen. titles}                                  & \textit{1.86 (0.00)}                                                                                   & \textit{\textbf{1.85 (0.00)}}          & \textit{1.75 (0.00)}              & \textit{0.15 (0.41)}                                                              & \textit{0.90 (0.22)}                                                                     & 1.82 (0.00)                                                                             & 1.80 (0.00)                                                                          \\ \hline
\end{tabular}
\caption{Debiasing results for gender across different debiasing methods of linear projection, Bolukbasi's hard debiasing and different variants of LPSG debiasing. We provide SEAT effect sizes with p-values of the test in parentheses. PCA for Linear Projection was done on gendered word pairs. IR stands for Information Retention.}
\label{tab:genderseatall}
\end{table*}

\begin{table*}
\small
\centering
\begin{tabular}{|c|c|c|ccc|}
\hline
\multirow{2}{*}{\textbf{\begin{tabular}[c]{@{}c@{}}Test \\ Type\end{tabular}}} & \multirow{2}{*}{\textbf{\begin{tabular}[c]{@{}c@{}}Description\\ (vs caste)\end{tabular}}} & \multirow{2}{*}{\textbf{\begin{tabular}[c]{@{}c@{}}Original \\ Score\end{tabular}}} & \multicolumn{3}{c|}{\textbf{Linear Projection}}                             \\ \cline{4-6} 
                                                                              &                                                                                            &                                                                                     & \textbf{ghasiya - desai} & \textbf{ghasiya - pandit} & \textbf{PCA}         \\ \hline
\multirow{2}{*}{\textbf{WEAT}}                                                 & occupations                                                                                & 1.44 (0.00)                                                                         & 1.34 (0.00)              & \textbf{0.78 (0.09)}      & 1.21 (0.02)          \\
                                                                              & adjectives                                                                                 & 1.52 (0.00)                                                                         & 1.51 (0.00)              & \textbf{1.31 (0.01)}      & 1.33 (0.00)          \\ \hline
\multirow{2}{*}{\textbf{SEAT}}                                                 & occupations                                                                                & 1.26 (0.00)                                                                         & 1.17 (0.00)              & \textbf{0.67 (0.00)}      & 0.89 (0.00)          \\
                                                                              & adjectives                                                                                 & 1.40 (0.00)                                                                         & 1.36 (0.00)              & 1.18 (0.00)               & \textbf{1.18 (0.00)} \\ \hline
\end{tabular}
\caption{Debiasing results for caste across different methods of choosing caste subspace. We provide WEAT and SEAT effect sizes with p-values of the test in parentheses. PCA was conducted on a list of caste names containing both upper and lower castes.}
\label{tab:castedebias}
\end{table*}

\begin{table*}
\small
\centering
\begin{tabular}{|c|c|c|c|ccc|c|}
\hline
\multirow{2}{*}{\textbf{\begin{tabular}[c]{@{}c@{}}Test \\ Type\end{tabular}}} & \multicolumn{2}{c|}{\multirow{2}{*}{\textbf{Description}}} & \multirow{2}{*}{\textbf{\begin{tabular}[c]{@{}c@{}}Original \\ Score\end{tabular}}} & \multicolumn{3}{c|}{\textbf{Linear Projection}}                                                                                                                                                          & \textbf{LPSG}                                                    \\ \cline{5-8} 
                                                                              & \multicolumn{2}{c|}{}                                      &                                                                                     & \textbf{\begin{tabular}[c]{@{}c@{}}Acharya - \\ Nasir\end{tabular}} & \textbf{\begin{tabular}[c]{@{}c@{}}PCA\\ entities\end{tabular}} & \textbf{\begin{tabular}[c]{@{}c@{}}PCA\\ lastnames\end{tabular}} & \textbf{\begin{tabular}[c]{@{}c@{}}w/o \\ entities\end{tabular}} \\ \hline
\multirow{3}{*}{\textbf{WEAT}}                                                 & \multicolumn{2}{c|}{adjectives vs religion terms}          & 1.28 (0.01)                                                                         & 1.28 (0.01)                                                         & 1.28 (0.01)                                                     & \textbf{0.91 (0.04)}                                             & 0.92 (0.06)                                                      \\
                                                                              & \multicolumn{2}{c|}{adjectives vs lastnames}               & 1.55 (0.00)                                                                         & 1.57 (0.00)                                                         & 1.55 (0.00)                                                     & 0.71 (0.10)                                                      & \textbf{0.71 (0.11)}                                             \\ \cline{2-8} 
                                                                              & \textit{IR}    & \textit{religious entities vs religion}   & \textit{1.75 (0.00)}                                                                & \textit{1.61 (0.00)}                                                & \textit{\textbf{1.72 (0.00)}}                                   & \textit{1.54 (0.00)}                                             & \textit{1.59 (0.00)}                                             \\ \hline
\multirow{3}{*}{\textbf{SEAT}}                                                 & \multicolumn{2}{c|}{adjectives vs religion terms}          & 1.20 (0.00)                                                                         & 1.22 (0.00)                                                         & 1.19 (0.00)                                                     & 0.85 (0.00)                                                      & \textbf{0.85 (0.00)}                                             \\
                                                                              & \multicolumn{2}{c|}{adjectives vs lastnames}               & 1.41 (0.00)                                                                         & 1.43 (0.00)                                                         & 1.41 (0.00)                                                     & 0.70 (0.00)                                                      & \textbf{0.68 (0.00)}                                             \\ \cline{2-8} 
                                                                              & \textit{IR}    & \textit{religious entities vs religion}   & \textit{1.69 (0.00)}                                                                & \textit{1.52 (0.00)}                                                & \textit{\textbf{1.65 (0.00)}}                                   & \textit{1.43 (0.00)}                                             & \textit{1.50 (0.00)}                                             \\ \hline
\end{tabular}
\caption{Debiasing results for religion across different methods of choosing religion subspace and LPSG. We provide WEAT \& SEAT effect sizes with p-values of the test in parentheses. We experimented with conducting PCA over a list of religious entities and over a list of religious lastnames.}
\label{tab:religiondebias}
\end{table*}

In Table \ref{tab:religiondebias}, we see that linear debiasing by conducting PCA over a list of religious entities is not able to debias much in any of the tests. The same could be observed for linear debiasing using single set of common last names \{$\vec{acharya}-\vec{nasir}$\}. However, if we linear debias by PCA over a list of lastnames, we are able to debias significantly. Although the Information Retention WEAT effect size is less than the previous methods, they did not even affect the religion bias which is our primary goal. Zhou's variant for religion debias performs well since it is able to debias competitively as well as retains greater amount of necessary religion information.  Refer to Appendix \ref{app:wordlists} for the word lists used in the test.

\begin{table*}
\centering
\small
\begin{tabular}{|l|l|l|}
\hline
\textbf{Description}                                                                                           & \textbf{Word list type}  & \textbf{List}                                                                                                                      \\ \hline
\multirow{4}{*}{\begin{tabular}[c]{@{}l@{}}Math, Arts vs \\ Gender specific \\ words\end{tabular}}             & Math words               & {[}ganit, beejganit, jyamiti, kalan, sameekaran, ganna, sankhya, yog{]}                                                            \\
                                                                                                               & Arts words               & {[}kavita, kala, nritya, sahitya, upanyas, raag, naatak, murti{]}                                                                  \\
                                                                                                               & Male gendered words      & {[}purush, aadmi, ladka, bhai, pati, chacha, maama, beta{]}                                                                        \\
                                                                                                               & Female gendered words    & {[}mahila, aurat, ladki, behen, patni, chachi, maami, beti{]}                                                                      \\ \hline
\multirow{4}{*}{\begin{tabular}[c]{@{}l@{}}Science, Arts vs \\ Gender specific \\ words\end{tabular}}          & Science terms            & {[}vigyan, praudyogiki, bhautik, rasayan, prayogshala, niyam, prayog, khagol{]}                                                    \\
                                                                                                               & Arts terms               & {[}kavita, kala, naach, nritya, sahitya, upanyas, raag, naatak{]}                                                                  \\
                                                                                                               & Male gendered words      & {[}bhai, chacha, daada, beta, purush, pati, aadmi, ladka{]}                                                                        \\
                                                                                                               & Female gendered words    & {[}behen, chachi, daadi, beti, mahila, patni, aurat, ladki{]}                                                                      \\ \hline
\multirow{4}{*}{\begin{tabular}[c]{@{}l@{}}Adjectives vs\\ Gender specific\\ words\end{tabular}}               & Stereo male adjectives   & {[}krodhit, shramik, takatwar, nipun, veer, sahsi, diler{]}                                                                        \\
                                                                                                               & Stereo female adjectives & {[}sundar, sharm, aakarshak, manmohak, madhur, gharelu, kamzor{]}                                                                  \\
                                                                                                               & Male gendered words      & {[}purush, aadmi, ladka, bhai, pati, chacha, maama, beta{]}                                                                        \\
                                                                                                               & Female gendered words    & {[}mahila, aurat, ladki, behen, patni, chachi, maami, beti{]}                                                                      \\ \hline
\multirow{4}{*}{\begin{tabular}[c]{@{}l@{}}Gendered verbs\\ vs Gender \\ specific words\end{tabular}}          & Male verbs               & {[}gaya, aaya, khelta, baitha, leta, rehta, deta, padhta{]}                                                                        \\
                                                                                                               & Female verbs             & {[}gayi, aayi, khelti, baithi, leti, rehti, deti, padhti{]}                                                                        \\
                                                                                                               & Male gendered words      & {[}purush, aadmi, ladka, bhai, pati, chacha, maama, beta{]}                                                                        \\
                                                                                                               & Female gendered words    & {[}mahila, aurat, ladki, behen, patni, chachi, maami, beti{]}                                                                      \\ \hline
\multirow{4}{*}{\begin{tabular}[c]{@{}l@{}}Gendered \\ adjectives vs \\ Gender \\ specific words\end{tabular}} & Male verbs               & {[}accha, bura, ganda, lamba, chota, meetha, neela, bada, pehla{]}                                                                 \\
                                                                                                               & Female verbs             & {[}acchi, buri, gandi, lambi, choti, meethi, neeli, badi, pehli{]}                                                                 \\
                                                                                                               & Male gendered words      & {[}purush, aadmi, ladka, bhai, pati, chacha, maama, beta{]}                                                                        \\
                                                                                                               & Female gendered words    & {[}mahila, aurat, ladki, behen, patni, chachi, maami, beti{]}                                                                      \\ \hline
\multirow{4}{*}{\begin{tabular}[c]{@{}l@{}}Gendered \\ titles vs\\ Gender specific\\ words\end{tabular}}       & Male titles              & {[}adhyapak, shishya, vidvan, saadhu, kavi, chhatr, pradhanacharya, mahoday{]}                                                     \\
                                                                                                               & Female titles            & \begin{tabular}[c]{@{}l@{}}{[}adhyapika, shishyaa, vidushi, saadhvi, kavitri, chhatra, pradhanacharya, \\ mahodaya{]}\end{tabular} \\
                                                                                                               & Male gendered words      & {[}purush, aadmi, ladka, bhai, pati, chacha, maama, beta{]}                                                                        \\
                                                                                                               & Female gendered words    & {[}mahila, aurat, ladki, behen, patni, chachi, maami, beti{]}                                                                      \\ \hline
\multirow{4}{*}{\begin{tabular}[c]{@{}l@{}}Gendered \\ entities vs\\ Gender specific\\ words\end{tabular}}     & Male entities            & {[}pajama, ghada, kurta, phool, kapda, pahiya, yantra, putla, taala{]}                                                             \\
                                                                                                               & Female entities          & {[}almaari, chadar, poshaak, bijli, buddhi, tasvir, ghadi, raakhi, kameez{]}                                                       \\
                                                                                                               & Male gendered words      & {[}purush, aadmi, ladka, bhai, pati, chacha, maama, beta{]}                                                                        \\
                                                                                                               & Female gendered words    & {[}mahila, aurat, ladki, behen, patni, chachi, maami, beti{]}                                                                      \\ \hline
\end{tabular}
\caption{Word lists for Gender WEAT and SEAT tests}
\label{tab:genderwordlists}
\end{table*}

\begin{table*}
\centering
\small
\begin{tabular}{|l|l|l|}
\hline
\textbf{Description}                                                                     & \textbf{Word List type}                 & \textbf{List}                                                                   \\ \hline
\multirow{4}{*}{\begin{tabular}[c]{@{}l@{}}Occupations\\ vs Caste\end{tabular}} & Stereo Upper caste occupations & {[}vyapar, jameendar, sunar, guru, munim, chikitsak, pandit{]}         \\
                                                                                & Stereo Lower caste occupations & {[}safai, dhobi, mallah, maali, naai, mochi, machuara{]}               \\
                                                                                & Upper caste names              & {[}thakur, brahmin, rajput, kshatriya, arya, jaat, baniya, kayastha{]} \\
                                                                                & Lower caste names              & {[}dalit, shudra, bhangi, chamaar, valimiki, harijan, chuhda, jatav{]} \\ \hline
\multirow{4}{*}{\begin{tabular}[c]{@{}l@{}}Adjectives\\ vs Caste\end{tabular}}  & Upper caste adjectives         & {[}ameer, gyani, veer, taakatvar, sundar, ucch, sahsi{]}               \\
                                                                                & Lower caste adjectives         & {[}neech, ghrana, ganda, kamzor, gareeb, agyani, nirbal{]}             \\
                                                                                & Upper caste names              & {[}thakur, brahmin, rajput, kshatriya, arya, jaat, baniya, kayastha{]} \\
                                                                                & Lowe caste names               & {[}dalit, shudra, bhangi, chamaar, valimiki, harijan, chuhda, jatav{]} \\ \hline
\end{tabular}
\caption{Word lists for Caste WEAT and SEAT tests}
\label{tab:castewordlists}
\end{table*}

\begin{table*}
\centering
\small
\begin{tabular}{|l|l|l|}
\hline
\textbf{Description}                                                                             & \textbf{Word List type} & \textbf{List}                                                                  \\ \hline
\multirow{4}{*}{\begin{tabular}[c]{@{}l@{}}Adjectives\\ vs\\ Religion \\ terms\end{tabular}}     & Positive adjectives     & {[}shikshit, veer, ucch, sahsi, shant, dayalu, safal{]}                        \\
                                                                                                 & Negative adjectives     & {[}neech, ghrana, ashikshit, hinsak, krodhi, nirdayi, atyachaari{]}            \\
                                                                                                 & Hindu religion terms    & {[}hindu, bhagwan, geeta, brahmin, pandit, mandir, ram, vrat{]}                \\
                                                                                                 & Muslim religion terms   & {[}musalman, allah, quran, shiya, sunni, masjid, muhammad, roza{]}             \\ \hline
\multirow{4}{*}{\begin{tabular}[c]{@{}l@{}}Adjectives\\ vs \\ Religion\\ Lastnames\end{tabular}} & Positive adjectives     & {[}shikshit, veer, ucch, sahsi, shant, dayalu, safal{]}                        \\
                                                                                                 & Negative adjectives     & {[}neech, ghrana, ashikshit, hinsak, krodhi, nirdayi, atyachaari{]}            \\
                                                                                                 & Hindu lastnames         & {[}sharma, verma, agrawal, gupta, chauhan, bansal, mittal, singh, chaudhary{]} \\
                                                                                                 & Muslim lastnames        & {[}yusuf, malik, khan, ansari, sheikh, abdullah, ahmad, pathan, mirza{]}       \\ \hline
\multirow{4}{*}{\begin{tabular}[c]{@{}l@{}}Religious\\ entities\\ vs Religion\end{tabular}}      & Hindu religion terms    & {[}bhagwan, geeta, brahmin, pandit, mandir, ram, vrat{]}                       \\
                                                                                                 & Muslim religion terms   & {[}allah, quran, shiya, sunni, masjid, muhammad, roza{]}                       \\
                                                                                                 & Hindu religion          & {[}hindu, hindutva{]}                                                          \\
                                                                                                 & Muslim religion         & {[}musalman, islam{]}                                                          \\ \hline
\end{tabular}
\caption{Word lists for Religion WEAT and SEAT tests}
\label{tab:religionwordlists}
\end{table*}

\begin{table*}
\small
\centering
\begin{tabular}{|c|c|c|}
\hline
\textbf{Description}                                                                                         & \textbf{Word List type} & \textbf{List}                                                                   \\ \hline
\multirow{4}{*}{\begin{tabular}[c]{@{}c@{}}Adjectives v/s \\ Rural and \\ Urban \\ Occupations\end{tabular}} & Positive Adjectives     & {[}ameer, gyani, veer, takatvar, sundar, ucchh, sahsi{]}                        \\
                                                                                                             & Negative Adjectives     & {[}neech, ganda, ghrana, kamzor, gareeb, agyani, nirbal{]}                      \\
                                                                                                             & Urban Occupations       & {[}banker, vyavsayi, engineer, vakeel, vaigyanik, chaalak, abhineta, manager{]} \\
                                                                                                             & Rural Occupations       & {[}lohar, jalvahak, kisaan, gwala, charwaaha, kumhar, jameendar, julaha{]}      \\ \hline
\end{tabular}
\caption{Word lists for Rural v/s Urban Occupations WEAT and SEAT tests}
\label{tab:rural-urban}
\end{table*}

\end{document}